\newcommand{\nvector}[2]{\ensuremath{\left({#1}_1, {#1}_2, \ldots, {#1}_{#2}\right)}}

\documentclass[times,10pt,twocolumn]{article} 
\usepackage{latex8}
\usepackage{times}

\usepackage{graphicx, hyperref, breakurl}

\begin{document}

\title{A Computational Approach to Style in American Poetry\\%
\small 2007 accepted manuscript of \url{https://doi.org/10.1109/ICDM.2007.76}}

\author{David M. Kaplan\\
Dept. of Computer Science \\ Princeton University\\ dkaplan@alumni.princeton.edu\\
\and
David M. Blei\\
Dept. of Computer Science \\ Princeton University\\ blei@cs.princeton.edu\\
}

\maketitle

\thispagestyle{empty}

% ABSTRACT
\begin{abstract}
We develop a quantitative method to assess the style of American
poems and to visualize a collection of poems in relation to one
another. Qualitative poetry criticism helped guide our development
of metrics that analyze various orthographic, syntactic, and
phonemic features. These features are used to discover comprehensive
stylistic information from a poem's multi-layered latent structure,
and to compute distances between poems in this space. Visualizations
provide ready access to the analytical components. We demonstrate
our method on several collections of poetry, showing that it
better delineates poetry style than the traditional word-occurrence
features that are used in typical text analysis algorithms. Our
method has potential applications to academic research of texts, to
research of the intuitive personal response to poetry, and to making
recommendations to readers based on their favorite poems.
\end{abstract}

% INTRODUCTION
\section{Introduction}
There is considerable ongoing research in natural language processing
to extract semantic content from prose texts.  The more mystical (if
less lucrative) realm of poetry, however, has gone largely unexplored.
The meek tradition of quantitative poetry analysis, dating back at
least 60 years to Josephine Miles's examination of frequent adjectives
by hand \cite{Miles46}, can profit from modern computational
techniques.  In particular, computers can give insight
into the latent structures that together form a poem's style. 
%CLD: Miles \cite{Miles57} proposed that examining relative part of speech
%CLD: frequencies would reveal a distinct ``phrasal'' or ``clausal'' style;
%CLD: others have studied formality \cite{Heylighen, Biber}, a stylistic
%CLD: dichotomy in Chinese poetry \cite{Li}, and determining the type of
%CLD: text such as ``poetry'' or ``drama'' \cite{Grzybek}.

Here, we attempt to computationally capture a comprehensive scope of
poetic style.  Further, we aim to make the results of the analysis
readily accessible through visualization of stylistic similarity among
poems in a collection.

This method has many potential applications, such as:~a personal
recommendation system based on style; academic exploration of how
particular poets differ from or influence one another; assessment of
how different elements affect readers' intuitive perception of a
poem's overall style; and how important style is compared to semantic
content in overall reader preference.

To achieve these goals, we embed each poem in a vector space that was
developed from an extensive survey of literary scholarship.  We then
use principal components analysis (PCA) to visualize collections.
With many poetry collections, our method showed success in a
variety of areas:~differentiating poems from poets with different
styles; verifying consistency within a single long poem; showing
evidence of known mentoring relationships; and more.  These results
were taken from both the colorful visual projections and the
statistical analysis of poem distances in the stylistic feature space.

In Section 2, we describe the approach that we took to map poems into
a quantitative vector space and visualize them.  Section 3 details analysis covering 81 poems by 18 poets, and a comparison to
traditional vector-space models.  Section 4 reviews the previous work
from various fields of study that laid the foundation for this
research.  Section 5 provides further discussion and possibilities for
the future.

\section{A Vector Space for Poetry}
Our first contribution is embedding poetry into a vector space for
analysis.  We focused on stylistic elements of poetry, foregoing
semantic content.  Our goal was to map any poem text to a
multi-dimensional vector that accurately represents the poem's place
in stylistic space.

Predominant approaches of modern text analysis focus on word
occurrence \cite{Manning, Baeza-Yates:1999}, which is not appropriate
to our goals. Word occurrence is used primarily to determine semantic
content, while we are concerned with style. Moreover, collapsing text
into a ``bag of words'' loses the structural information that is
critical to style.  Diction per se can certainly have a stylistic
element, for instance affecting the formality of tone by choosing the
word ``egregious'' over ``very bad,'' but poetic style mostly derives
from relationships among words on multiple levels.

Instead, we identified different features that comprise a po\-em's
style and subsequently implemented them as computational functions of
the text.  The computation of these metrics maps the poem text to a
high-dimensional vector, with the computed value of each metric
providing the coordinate location in the corresponding
dimension:~where $f_i(p)$ for $(1 \leq i \leq N)$ are the metrics
taking text $p$ as input and producing scalar values, $p \mapsto
\nvector{f}{N}$.

\subsection{Features of style} \label{FeaturesSection}
A mix of ideas from the existing literature
of poetry criticism and personal intuition informed our decisions regarding
which features to consider. In total, we had 84 metric dimensions available, each falling under one of the features discussed here.  Features are
divided into three categories:~orthographic, based on the
letters or words as written (without higher-level interpretation);
syntactic, based on word function; and phonemic, based on sound.

\paragraph{Orthographic}
%The features of poems most readily available to a computer are
%orthographic.  These features are commonly used when discussing
%computational analysis of text, whether in daily life with the word
%count tool in Microsoft Word, in sentence length
%and similar histograms in statistical natural language processing
%textbooks \cite{Manning}, or in poetic analysis
%wherein, Miles states, ``The poetic line is the unit of measure'' \cite{Miles46}.

Our orthographic features were motivated by intuition and Miles's statement, ``The
poetic line is the unit of measure'' \cite{Miles46}.  The primary
features that we analyze are word count, number of
lines, number of stanzas, average line length (in words), average word
length, and average number of lines per stanza.  We also calculate
the frequencies of the most frequent noun, adjective, and verb
(respectively) in each poem as proxies for repetition.

\paragraph{Syntactic}
The frequencies of parts of speech (POS) reflect a poet's mode of
discourse.  Miles \cite{Miles67} examined the
adjective-noun-verb-connective (A-N-V-C) ratio for notable
English-language poets.  Miles \cite{Miles57} also examined phrasal
versus clausal type, a distinction partly manifested in POS
frequencies:~phrasal type has an ``abundance of adjectives and nouns,
in heavy modifications and compounding of subjects, in a variety of
phrasal constructions, including verbs turned to participles,'' while
clausal type has more ``relative and adverbial clauses, action [i.e.,
tensed verbs].''  Miles also discussed the dichotomy of ``adjectival''
versus ``predicative'' style, where predicative manifests itself in
``the dominance of verb over adjective'' \cite{Miles67}.

Heylighen and Dewale \cite{Heylighen} found that POS frequencies reflect the level of formality in different languages including English.
%They wrote in their study, ``Nouns, adjectives, articles and
%prepositions are more frequent in formal styles; pronouns, adverbs,
%verbs and interjections are more frequent in informal styles.''
Biber \cite{Biber} cites contractions as reflecting formality,
stating, ``contractions and first and second person pronouns share a
colloquial, informal flavor.''

We include both frequencies of contractions and of parts of speech aggregated to
different levels of specificity, e.g.~pronoun as well as first person
singular pronoun.

\paragraph{Phonemic}
Sound is vital to the experience of a poem.  Miles \cite{Miles46}
wrote, ``All patterns of repetition in sound [including]
assonance\ldots provide indeed some basis in the poetic material.''
Repetition is useful in poetry especially with sound, and the major
poetic sound devices are all analyzed.

Rhyme is the most well-known feature of poetry, prominent in 
nursery rhymes that children hear growing up.  Much modern poetry,
though, abandons a formal rhyme scheme, if not rhyme altogether.  This
actually increases the potential explanatory power of rhyme frequency
since its use is voluntary and subsequently more varied by poet.
There are different types of end rhyme, too, which we define
as:~identity rhyme, identical phoneme sequences; perfect rhyme, the
same phoneme sequence from the ultimate stressed vowel onward, but
differing in the preceding consonant; semirhyme, a perfect rhyme where
one word has an additional syllable at the end, such as ``stick'' and
``picket''; and slant rhyme, either identical ultimate stressed vowels
or phoneme sequences following the ultimate stressed vowel, but not
both.  All four types of rhyme and certain combinations thereof are considered
as features.

The next most prominent sound devices used in poetry are alliteration
(repetition of consonant sounds beginning words), assonance
(repetition of vowel sounds), and consonance (repetition of consonant
sounds), all of which are features that we compute.

\subsection{Computation of metrics}
We implemented each of the features above as a computational metric,
mapping each poem text into a feature vector, as described above.  We
set weights for each metric by which the raw value is multiplied, with
a setting of zero effectively turning off the metric.  Thus we could
cause certain metrics to contribute relatively more towards the total
stylistic distance between poems, either to reflect a personal sense
of the relative importance of metrics or to focus on specific features
of style.

% DMB: we said something like this earlier.
% DMK: true

% Each metric function translates the poem text into a numeric value.
% Applying all metric functions $f_i(p)$ for $(1 \leq i \leq N)$ maps
% any poem text $p$ to high-dimensional feature space:~$p \mapsto
% \nvector{f}{N}$.

To determine parts of speech, we used a rule-based POS tagger based on
\cite{Hepple}.  It was acquired already trained (on a \textit{Wall Street Journal} corpus).  While POS tagging is an area for
improvement, our comparison of the tagger's results with manual
tagging of a few real poems showed enough accuracy to produce
meaningful results.  We used the CMU Pronouncing Dictionary for North American English to translate words into phoneme
sequences for analysis of sound devices.

\subsection{Visualization with PCA}

The high-dimensional poem vectors are projected onto two dimensions to
present as accurate a depiction of relative poem similarity as
possible.  We used PCA, which reduces dimensionality while
preserving the greatest variance in the data.  The complexity of PCA
is determined by the singular value decomposition (SVD).  It took 5.7
minutes to run SVD with 50,000 poems and 11.3 minutes for 100,000
poems; a more realistically sized sample of 80 poems took under 0.05
seconds.  All runs were performed on a 1.80GHz Pentium 4 with 512MB
RAM.  Presentational accuracy decreases with more poems; a calculation
of ``stress,'' or reconstruction error, is available to indicate the
overall correctness of the visualization.

\section{Results}
In this section, we describe our analyses of poems by prominent
American poets covering a variety of periods and styles.\footnote{The
  full source and data are freely available from the first author for
  non-commercial use.}  The poets were selected from several sources,
including the \textit{Oxford Anthology of Modern American Poetry}
\cite{OxfordAnthology} and Wikipedia \cite{PoetryoftheUS}.

% DMB: we say most of this later.  we don't need to say: "we did more than
% we are saying here..."
% DMK: true, ok

% As a consequence of the novelty of this research, there exists no
% comprehensive, quantitative baseline with which to compare the results
% produced here.  Consequently, we took many different approaches to
% assess our method's performance, such as looking for a quantitative
% reflection of statements from qualitative academic poetry analysis,
% and examining clustering of poems by poet.  We also compared our
% method to a bag of words cosine distance algorithm, using poet
% clustering results.  Space limitations preclude discussion of every
% analysis of every collection.

We set the weights trying to ensure that no individual metric(s)
drowned out the explanatory power of the others.
This was admittedly ad hoc, and can be improved/automated in the future.
We used the same weights for all analyses.

% DMB: removed for space
% DMK: put back in
% This was ad hoc; automation of weight-setting is an area of future
% work.

\subsection{Feature analysis}
\label{sec:featureanalysis}
To give a better sense of the full computational process, we provide
below excerpts from three poems along with their computed values for a
few salient feature metrics:~frequencies of perfect end rhyme, first
person singular pronouns, and coordinating conjunctions.  
These feature values reflect the poems' overall relationships, which are in turn reflected in the
visualization.

First, the opening five lines of the famous ``The Road Not Taken'' by
Robert Frost are: ``Two roads diverged in a yellow wood, / And sorry I
could not travel both / And be one traveler, long I stood / And looked
down one as far as I could / To where it bent in the undergrowth.''
Here, the perfect end rhyme of wood/stood/could and both/growth
figures prominently.  Frost rhymes not only pairs of lines but
triplets; he follows this \textit{abaab} rhyme scheme throughout the
20-line poem.  The first person perspective is also noteworthy; these
are not objective descriptions but subjective musings.  That three
lines begin with ``And'' is perhaps the first feature noticed by a
quick glance; this use of coordinating conjunctions continues through
the poem.

Second, the opening two stanzas of Louise Gluck's more modern
``Parable of Faith'' are: ``Now, in twilight, on the palace steps /
the king asks forgiveness of his lady.~/ / He is not / duplicitous; he
has tried to be / true to the moment; is there another way of being /
true to the self?''  In contrast to Frost, Gluck has no formal rhyme
scheme.  Also unlike Frost, Gluck writes in third person (appropriate
for a ``Parable'').  Finally, there are no conjunctions here.

Third, the closing sestet of a sonnet by Edna St.~Vincent Millay, entitled ``Love Is Not All,'' is:
``It well may be that in a difficult hour, /
Pinned down by pain and moaning for release, /
Or nagged by want past resolution's power, /
I might be driven to sell your love for peace, /
Or trade the memory of this night for food.~/
It well may be.~I do not think I would.''
Millay's work is between Frost's and Gluck's.  She follows the traditional English sonnet form, yielding plenty of perfect end rhymes, though only two lines per rhyme to Frost's three (or two), and the final couplet has a slant rhyme instead; her poem's perfect rhyme metric value is 0.139, to Frost's 0.278 and Gluck's zero.  Although not shown by the excerpt, Millay's opening octave is (traditionally) an objective setup of the suggestion that ``Love Is Not All.''  
While she comes around to a first person statement in the final twist, this is balanced by a complete lack of first person in the preceding lines; her first person pronoun metric value is 0.032, to Frost's 0.063 and Gluck's zero.  
Millay's poem flows differently than Gluck's, partly from extensive use of coordinating conjunctions and partly from the sonnet's iambic meter, which Frost uses less strictly.
Millay's coordinating conjunction metric value is 0.104 and Frost's 0.063, against Gluck's zero.

% DMK: embedded in text (above)
% \begin{table}[t]
% \centering
% \begin{tabular}{|l|r|r|r|} \hline
% \textbf{Poet}&\textbf{Perf. Rhyme}&\textbf{1S Pron.}&\textbf{Co. Conj.}\\ \hline
% Frost & 0.278 & 0.063 & 0.063\\ \hline
% Gluck & 0.000 & 0.000 & 0.000\\ \hline
% Millay & 0.139 & 0.032 & 0.104\\ \hline
% \end{tabular}
% \caption{Selected computed feature values of three poems discussed in Section~\ref{sec:featureanalysis}.}
% \label{FrostGluckMillayTable}
% \end{table}

PCA places Millay's poem about in between Frost's and Gluck's; see
Fig.~\ref{OxfordAnthology} (Frost's is 6, Millay's 37, and Gluck's 43).  
This reflects their overall computed
distances:~Frost and Gluck are the farthest apart, at 234.6, while
Millay is almost equidistant from both, 182.9 from Frost and 178.5
from Gluck.

% DMK: condensed legend to save space
\begin{figure}[t]
\centering
% in full page width version, width was "0.495\textwidth"
  \includegraphics[width=\hsize]{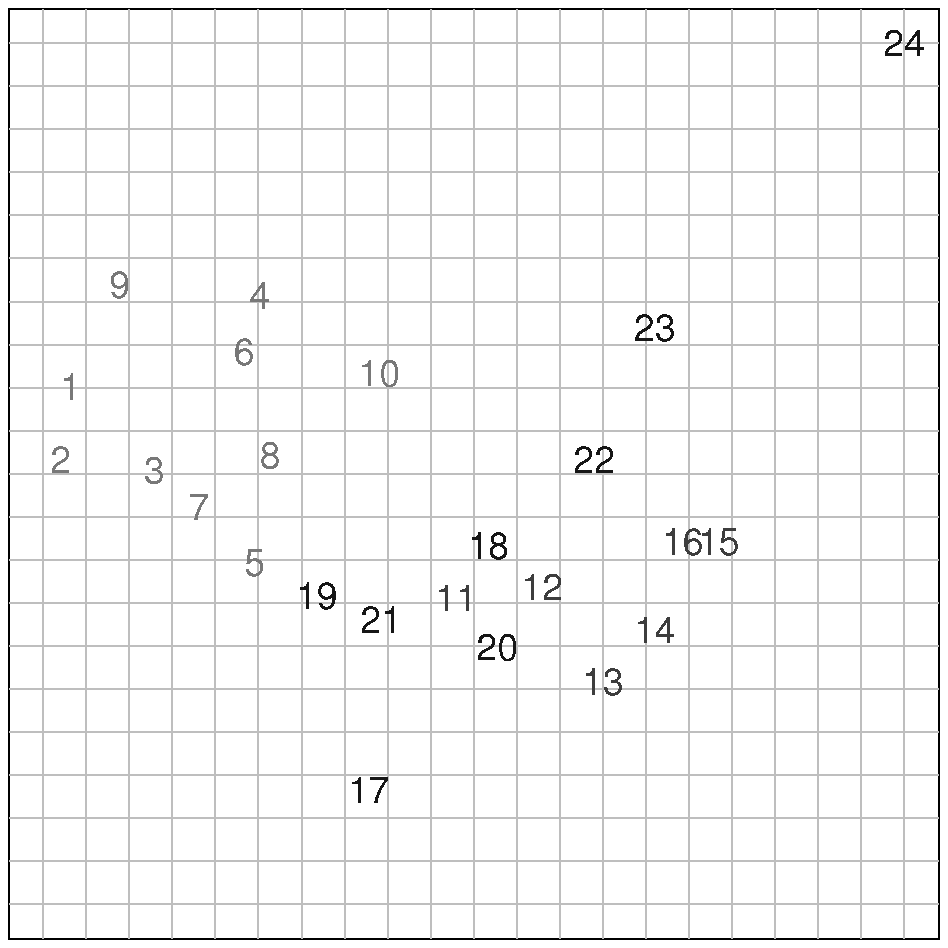} %\\MooreFrostOharaPlot.eps is COLOR, Zamzar_MFOgrey.ps
\begin{tabular}[t]{c}
  Legend:~1-10, Moore; 11-16, Frost; 17-24, O'Hara.
\end{tabular}
\caption{Similarity among poems from Moore, Frost, and O'Hara} \label{FrostMooreOhara}
\end{figure}

\subsection{Sample analysis}
We compared selected poems from
Robert Frost's \textit{North of Boston} (1915), poems
written by Marianne Moore, and selections from Frank O'Hara's famous \textit{Lunch Poems} (1964).  As the plot shows
(Fig.~\ref{FrostMooreOhara}), our method identifies a division
between the styles of these poets and represents this difference
visually.  O'Hara's work lies ``in between'' Frost's and Moore's; or, at least, O'Hara's poems are more similar to Frost's and Moore's, respectively, than are Frost's and Moore's to each other.  From the plot, O'Hara's ``Song (Is it dirty)'' (24) appears to be an outlier; this is supported by the computed values.

\begin{figure}[t]
\centering
  \includegraphics[width=\hsize]{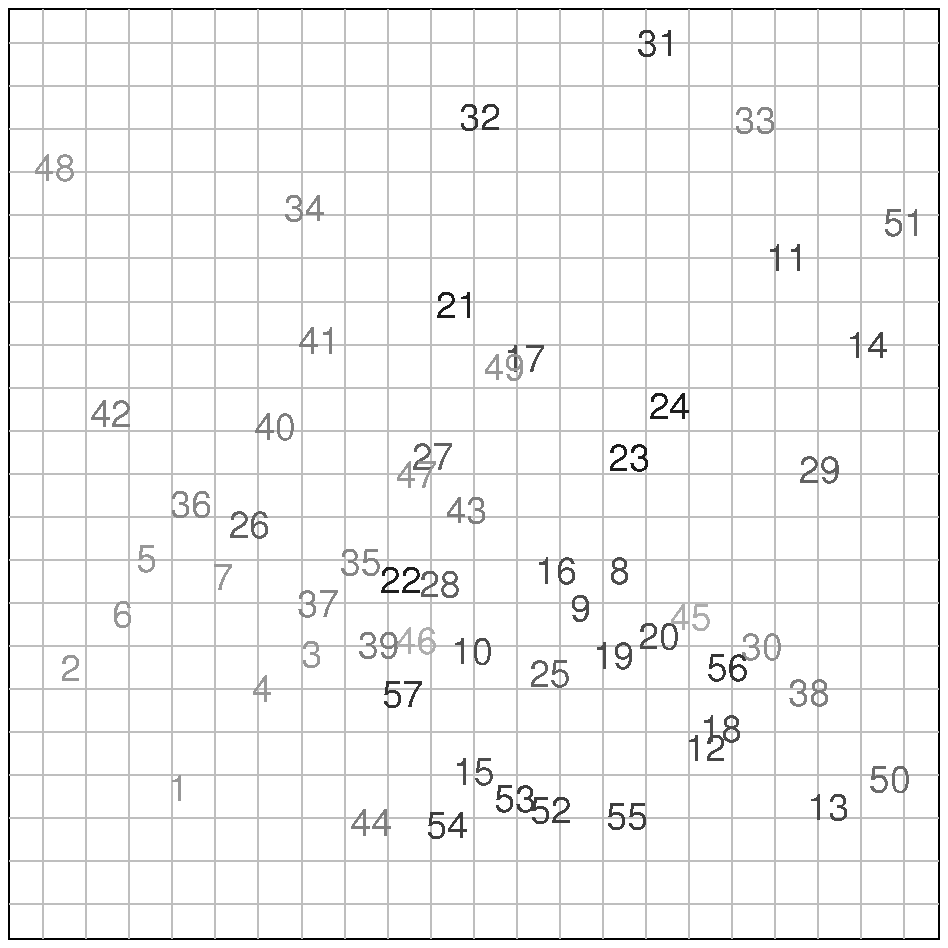} %\\OxfordPlot.eps is color
\begin{tabular}[t]{p{0.95\hsize}}
  Legend:~1-7, Frost; 8-10, Whitman; 11-14, Williams; 15-20, Stevens;
  21-24, Sexton; 25-29, Plath; 30, Pinsky; 31-32, Pound; 33-37,
  Millay; 38, Ginsberg; 39-44, Gluck; 45-46, Eliot; 47-49, Dickinson;
  50-51, Cummings; 52-55, Bishop; 56-57, Smith.
\end{tabular}
\caption{Similarity among poems from the \textit{Oxford
    Anthology of Modern American Poetry}.}
\label{OxfordAnthology}
\end{figure}

\begin{figure}
\centering
%DEPRECATED\psfig{file=oxfordanthologymean.ps, width=\hsize}
\includegraphics[width=\hsize]{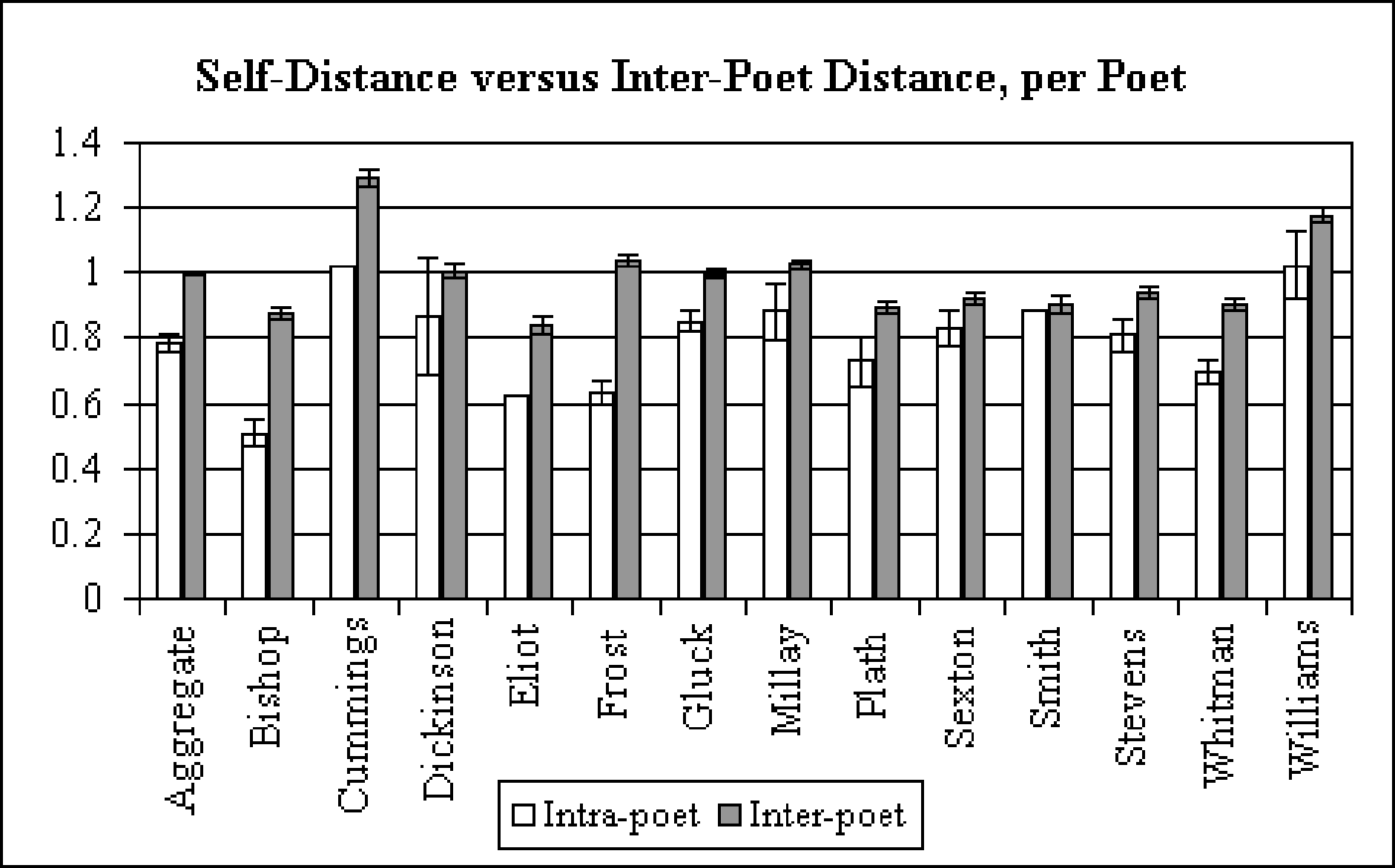}
\caption{Comparison of intra- and inter-poet distances for selections from the \textit{Oxford Anthology of Modern American Poetry}, using Euclidean distances calculated from the features described in Section \ref{FeaturesSection}.}
\label{OxfordAnthologyMean}
\end{figure}

While using a limited number of poets simplifies the relationships
present in the data and tends to result in more accurate
visualization, using a larger number of poets still yields significant
results. The works of 16 poets from the \textit{Oxford Anthology of
Modern American Poetry} \cite{OxfordAnthology} (plus Tracy K.~Smith)
yielded Fig.~\ref{OxfordAnthology}.  As expected, the accuracy of the
visualization drops (the display ``stress'' increases by 
50\%), but much clustering can still be seen.  For instance, the
Robert Frost poems (1-7) are grouped in the lower-left area; Louise
Gluck's three ``Circe'' poems (40-42) lie in the area above Frost;
Walt Whitman's three poems (8-10) sit slightly below and right of
center; and the four poems from Elizabeth Bishop (52-55) are adjacent
in the middle of the bottom edge of the screen.
% DMK: condense legend to save space
% New: intro self-distance/inter-dist here.
From the true, higher-dimensional distance data, we constructed a
chart (Fig.~\ref{OxfordAnthologyMean}) showing the observed degree of
clustering by poet.  The first data series (white columns) shows the
mean ``self-distance'' for each poet and in aggregate, calculated by
taking the mean of the Euclidean distances of all poem pairs for a
given poet.  The second data series (gray columns) shows the mean
``inter-poet'' distance for each poet and in aggregate.  Here, all
pairs containing one and only one poem of a poet are considered when
calculating mean distance.
% DMB: removed these examples for space; the description of the
% measurements is clear.
% For example, there are six Wallace Stevens poems, and thus ${6
%   \choose 2} = \frac{(6)(5)}{2} = 15$ Stevens poem pairs; the mean
% of the 15 distances is his mean self-distance.  The error bars shown
% are standard errors of the mean.
% For example, there are 51 poems not by Stevens, and six Stevens poems,
% yielding $(51)(6) = 306$ poem pairs with one and only one Stevens
% poem; the mean Stevens inter-poet distance is the mean of these 306
% distances.
All poets (and the aggregate) showed smaller mean intra-poet distances
than inter-poet distances.
The short length of both Dickinson's and Williams's poems may account for their high
variance.  From the visualization and the statistical analysis,
Elizabeth Bishop, Robert Frost, and Walt Whitman seem to have the most
consistent styles.  The chart shows values normalized by a scalar
factor, such that mean aggregate inter-poet distance is 1, and standard
error of the mean error bars.

% DMB removed
%The difference between aggregated intra-poet and inter-poet distances is
%significant, reflecting our method's ability to distinguish
%similar and different styles.  The mean (not normalized) distance between poems from
%distinct poets was 0.1865, compared to 0.1465 for two poems from the
%same poet.  Distances of 0.1 or smaller comprised over 16\% of the
%intra-poet distances but only 2.3\% of the inter-poet distances,
%while distances over 0.25 contributed almost 15\% of the inter-poet
%distances but under 3\% of the intra-poet distances.  This means
%that randomly selected poem pairs with distances in the ``extremes'' (lower
%10\% and upper 10\%) could be classified as inter-poet or intra-poet
%with a high degree of certainty.

\subsection{Comparison with word occurrence}
As a quantitative gauge of performance, we implemented a bag of words cosine distance algorithm using term frequency (TF) and inverse document frequency (IDF) weights \cite{Baeza-Yates:1999}.  
A bag of words approach, though not stylistically oriented, provides a modern standard for document similarity.  
We applied it to the two sets of poetry analyzed above plus a set of only Moore and Frost, once with just TF weights and once with TF and IDF for each set.  
While our method calculates a lower intra-poet than inter-poet distance for every poet, the bag of words approach does not; 
when it does, our method usually reports a more statistically significant differential.

For the following two examples, we examined the difference of inter- and intra-poet distance.
To compare fairly, we scaled this value by the standard error of the mean for the inter-poet distance (per poet).  Thus, an algorithm---primarily TF-IDF is affected---is not penalized for producing more similar absolute values if the differentials are just as statistically significant.  The second example shows aggregates across all poets within the selections indicated, normalized the same way.  In both cases, bigger bars indicate more ability to differentiate style by poet.

\begin{figure*}
\centering
\begin{tabular}{cc}
\includegraphics[width=0.45\hsize]{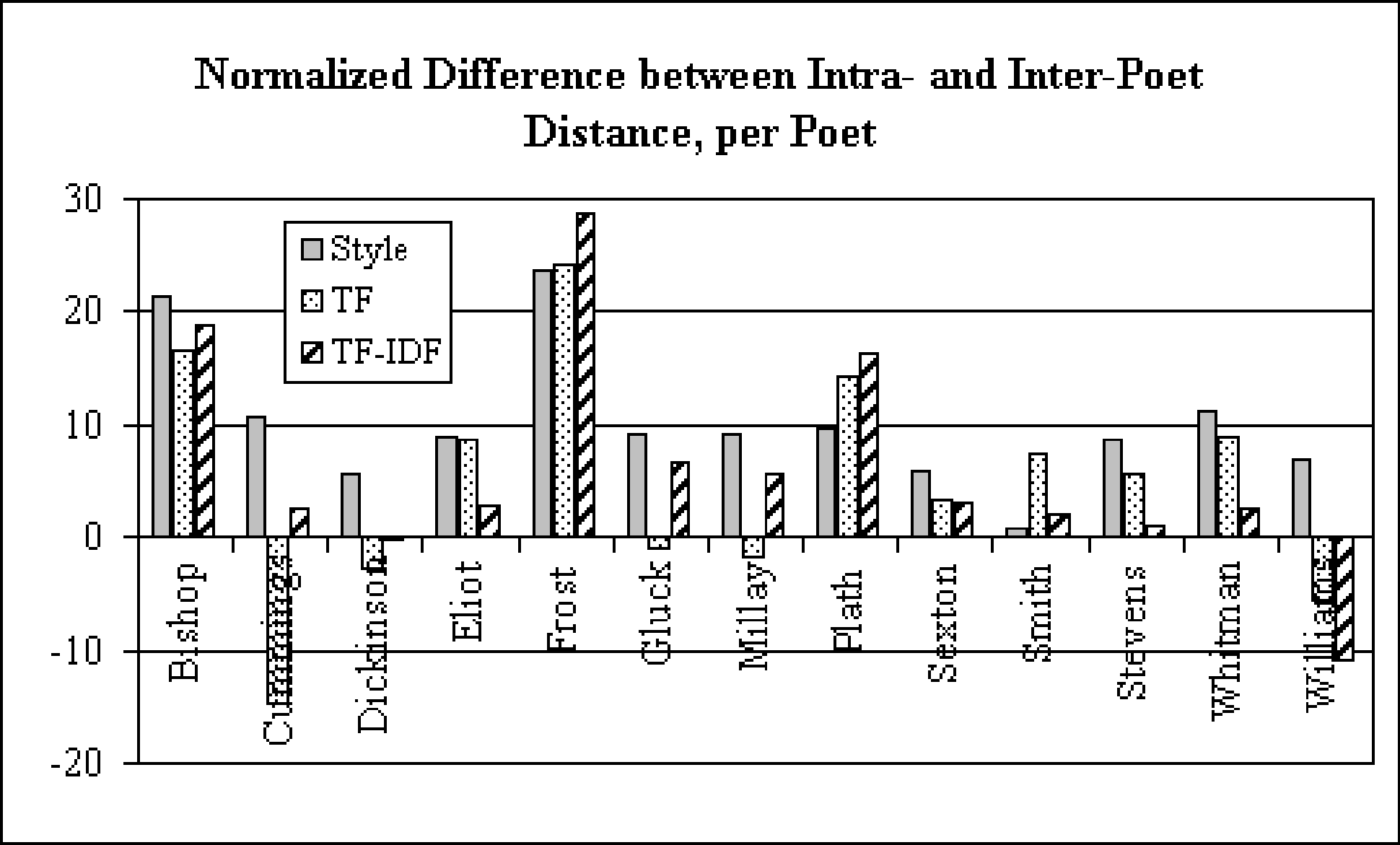} &
\includegraphics[width=0.45\hsize]{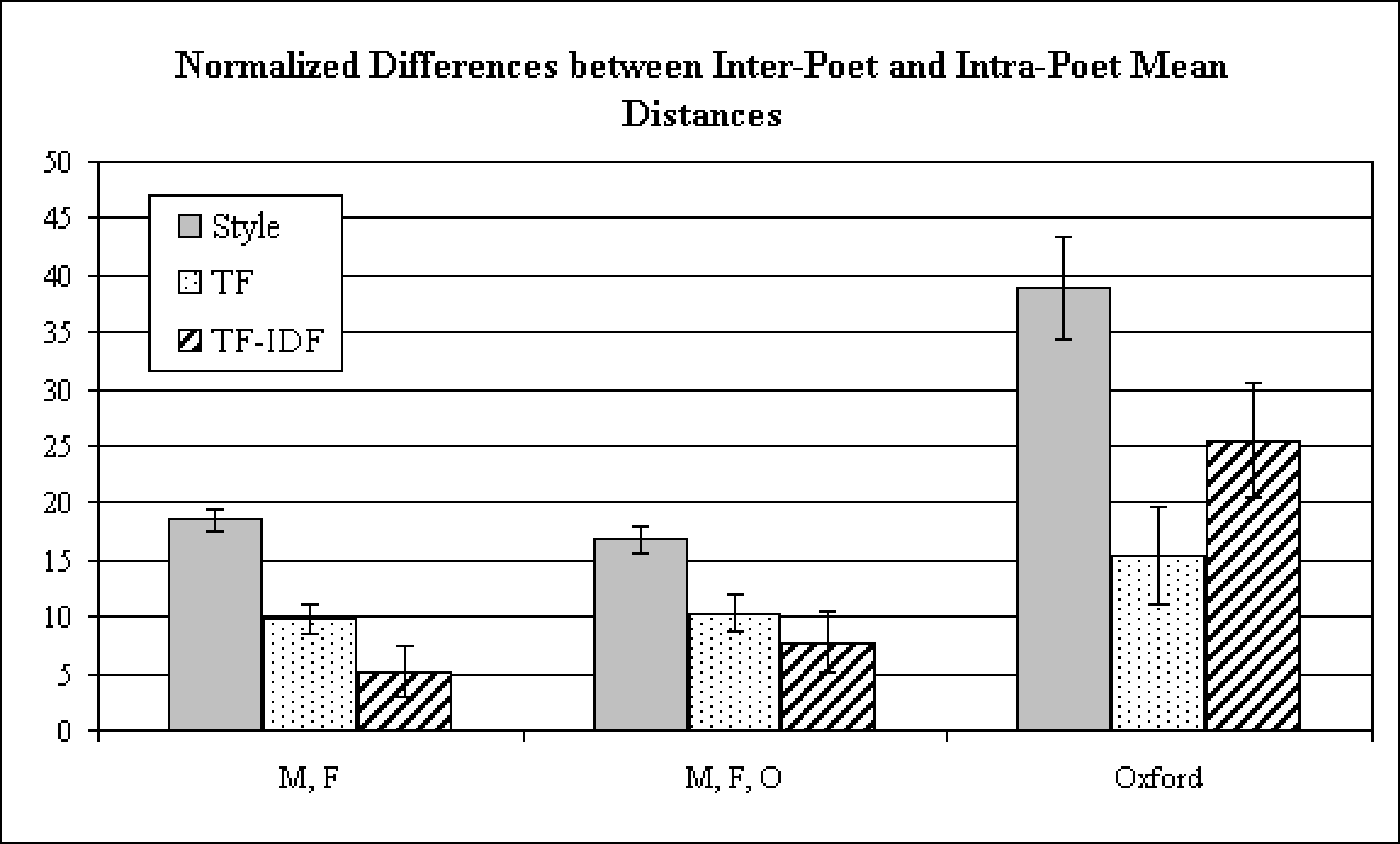}
\end{tabular}
\caption{Comparison to term frequencies.  Our algorithm is ``Style'';
  larger values indicate clustering by poet. (Left) \textit{Oxford
    Anthology} selections. (Right) Three
  collections.} \label{ComparisonOxf}
\end{figure*}

Comparing performance by poet in the \textit{Oxford Anthology}
selection of poems (Fig.~\ref{ComparisonOxf} Left), there are a few
instances where the bag of words algorithm (``TF'' or ``TF-IDF'')
performs better than ours (``Style'').  For these poets, perhaps the
algorithm is picking up the plethora of personal pronouns in Frost, or
the ``bee'' themed words in Plath's ``The Bee Meeting,'' ``Stings,''
``The Swarm,'' and ``The Arrival of the Bee Box.''  Overall, our
technique performed better, showing greater differentials in 10 of the
13 poets.

Comparing aggregate performance on three poetry selections (Moore and
Frost; Moore, Frost, and O'Hara; and the \textit{Oxford
  Anthology}-based collection), the advantage of our method over the
bag of words algorithm is significant (Fig.~\ref{ComparisonOxf}
Right).  This provides further quantitative evidence that our method
indeed captures latent stylistic features of poems, enough to discover
poem clustering by poet better than an existing method.

\subsection{Additional analysis}
The preceding examples were drawn from a much larger set of
explorations that included comparing well-known atypical poems such as
Whitman's ``O Captain! My Captain!'' with their more typical
counterparts and comparing sections within one long poem (e.g.,
Whitman's ``Song of Myself'') to check its consistency.  Additionally,
we ran sensitivity tests to ensure that metrics other than parts of
speech and length still showed poet clustering; this was especially
encouraging since the remaining metrics (rhyme, etc.)~are arguably the
least similar to the word frequency used in most current quantitative
textual analysis.

% DMB: !!! below, cite the book properly with bibtex.
% DMK: done

We also examined known poet mentoring relationships, particularly those of 
Williams and Creeley, Stevens and Ashbery, and Moore and Bishop. 
These were identified by L.~Keller 
in her book \textit{Re-Making It New:~Contemporary American Poetry and
the Modernist Tradition} \cite{Keller}, 
although she singled out Creeley-Williams as more personal support than stylistic influence.
Of 55 poet pairs that we examined---all
combinations of Ashbery, Bishop, Creeley, Dickinson, Frost, Ginsberg,
Gluck, Millay, Moore, Stevens, and Williams---the mean inter-poet
distance was 94 (SD=25).  The
Creeley-Williams distance was 104, while the other two pairs picked
out by Keller were among the closest:~Ashbery-Stevens was 50, and Bishop-Moore 59.

% DMB: (below) this is interesting, but not essential to our argument

% As a book review \cite{Willis} described,
% ``Keller argues that the younger poets, as they began their careers,
% were strongly influenced by their elders.''  As a particular example,
% ``It is safe to say that Bishop trained her own eye in Moore's
% fashion, becoming, as Keller states, a descriptive poet like Moore.''
% However, it is especially notable that the Creeley/Williams
% relationship was said to be ``more for personal reinforcement than for
% development of technique,'' since no particular similarity in their
% styles was found.

 Interestingly, poems from Williams's \textit{Spring and All} (1923)
 were very close (distance 54) to his other volume, \textit{The Wedge} (1944),
 when merging poems from each volume into a single
 ``poem.''  This suggests that the variability seen in his individual 
 poems may be largely due to their short length, while collectively the poems
 may gravitate around a consistent Williams style.

\section{Previous Work}
The two main contributors to statistical poetry analysis are Josephine
Miles and Marina Tarlinskaja, both of whom had to collect data by
hand.  Miles examines frequently used adjectives in English-language
poetry and studies adjective-noun-verb-connective proportions across
different eras~\cite{Miles46,Miles57,Miles67}.  Tarlinskaja statistically analyzes
poetry across many languages, specializing in meter and prosody \cite{Tarlinskaja, Tarlinskaja87}.

There has been less research in automated poetry analysis.
Contributions include the use of a connectionist model \cite{Hayward},
a Chinese poem classifier \cite{Li}, determination of Slavic textual
genre by word length distribution \cite{Grzybek}, and the prevalent
use of word frequencies \cite{Klarreich}.

\section{Discussion}

Our method was able to distinguish poetry texts based on a combination
of features not traditionally used in prose text analysis but
traditionally relied upon for poetry analysis.  We examined clustering
by poet as one proxy for performance, assuming that poets have
relatively consistent styles.  The results, both statistical and
visual, show that our method can discern stylistic similarity in
poetry.  Further, this ability is significantly greater than that of a
bag of words cosine distance algorithm with TF-IDF weights.

One area of future work is to explore other features of poetry.  The
one major stylistic area that went altogether untouched is
pro\-so\-dy, the rhythmic variations in stress and intonation, which
offers many challenges.
%Hurdles to proper prosody parsing include word disambiguation, intended unconventional stress patterns based on meter, and the malleability of monosyllabic words' stress from rhythmic and syntactic context \cite{Tarlinskaja}. 
Great progress has been made in text-to-speech algorithms for prose,
but their accuracy in poetic prosody is unknown.  Another area left
unexplored is visual style, important to such notable poets as ee
cummings.  Additional metrics that we did not use include additional
amalgamated metrics such as a full stop frequency; additional measures
of length; and a breakdown of verbs into transitive, intransitive, and
copula (linking).

A second area of future work is to look into other methods of
dimensionality reduction.  Possibilities include non-negative matrix
factorization \cite{Lee99} and variants of topic models such as the
author-topic model \cite{Baeza-Yates:1999}.

We developed a computational method of feature analysis for poetry,
guided by traditional qualitative and quantitative approaches.  We
enhanced this analytical engine with visualization and a graphical
interface.  Our analyses demonstrate considerable potential for this
approach.

%   We hope that this work
% will encourage further study of the computational analysis and
% visualized comparison of style in poetry.

\bibliographystyle{latex8}

 % \bibliography{dkaplan}

\end{document}